\documentclass{article}

\usepackage[dblblindworkshop, final]{neurips_2026}

\workshoptitle{Foundations of Language Model Security}

\usepackage[utf8]{inputenc} 
\usepackage[T1]{fontenc}    
\usepackage{hyperref}       
\usepackage{url}            
\usepackage{booktabs}       
\usepackage{amsfonts}       
\usepackage{nicefrac}       
\usepackage{microtype}      
\usepackage{xcolor}         %
\usepackage{amsmath}
\usepackage{graphicx}
\usepackage{float}
\usepackage{longtable} 
\usepackage{enumitem}
\setlist[itemize]{topsep=3pt, itemsep=1pt, parsep=0pt}
\setlist[enumerate]{topsep=3pt, itemsep=1pt, parsep=0pt}

\title{Global Security Blind Spots: Translation as a Safety Bypass in Low-Resource Languages}

\author{%
  \bf Jalil Alva \quad David Harper \quad Abigail E. Thornton \quad Rafael S. Villalobos Alva \\
  \bf Leonardo Kenji Minemura Suazo \quad Konstantinos Karageorgos \quad Diego Velazquez \\
  \bf Sergio A. Serrano \quad Marvin Vargas Flores \quad Aaron Schliem \quad Fernando F. Migone \\[0.6em]
  Welo Data
}

\begin{document}

\maketitle

\begin{abstract}
Large language models are deployed globally, yet safety alignment and evaluation remain concentrated in English: harmful requests refused in English can elicit unsafe completions when translated into low-resource languages. We evaluate ten widely deployed LLMs across 79 languages (English plus 78 low-resource) under a translation-bypass threat model, using a metric conditioned on English safety, which separates translation bypass from unsafety already present in English. English safety does not transfer reliably: conditioned on safe English completions, mean bypass rates range from 21.9\% to 44.4\%. Bypass risk concentrates in a subset of language families, a pattern broadly shared across models. The shared ordering is confounded with translation degeneracy, but its variation across models is significant and survives every control we apply. We also show that per-model bypass rates do not identify a stable ranking: nine of ten models change rank across three conditioning schemes that estimate the same quantity, with no change to any underlying response.
\end{abstract}

\vspace{-5pt}

\section{Introduction}

\subsection{Why multilingual safety is a security problem}

Large language models are deployed globally, but their safety evaluations and alignment remain centered on English \citep{yong2025state}. Guardrails that appear robust under English-only testing may fail when the same intent arrives in another language, and adversaries can exploit this through low-effort translation \citep{yong2023lowresource, deng2024multilingual, alghanim2024jailbreaking}. This follows a wider pattern: benchmarks such as Global-MMLU, MMLU-ProX, and XCOPA show that models perform best in English and get worse in lower-resource languages \citep{singh2025globalmmlu, xuan2025mmlupro, ponti2020xcopa}. Because safety relies on the same language understanding as everything else, alignment trained primarily in English may generalize poorly in the languages with the least coverage \citep{wang2024all}. Reporting habits make it worse: of 24 top-ranked models that publish system reports, 20 advertise broad multilingual support but only 5 report any multilingual safety red-teaming \citep{yong2025state}.

\subsection{Threat model: translation as a low-effort bypass}

This gap is not just a coverage problem, but something an attacker can use. We frame it with a simple threat model:

\begin{itemize}
  \item \textbf{Attacker goal.} Get a harmful completion with as little effort as possible.
  \item \textbf{Attacker capability.} Translate a prompt into many languages and query the model repeatedly, including trying several low-resource languages.
  \item \textbf{Attack strategy.} A two-step approach: probe in English, if the English response is safe, translate the same intent into one or more low-resource languages and retry. The attack succeeds if any translated prompt produces a harmful completion.
\end{itemize}

This takes no jailbreak skill. What matters is not whether a model ever slips up outside English, but whether it fails widely and reliably across translations of prompts it handled safely in English. That is what our metric captures, and what an unconditional (overall) unsafe rate blurs together.

\subsection{Contributions}

Prior work established that translation bypass happens, and we do not claim to have found it. Our contribution is about measurement, and it has three parts: (i) a metric that separates translation bypass from unsafety that already existed in English, (ii) a formal test of whether the risk is organized by language type, run across 78 low-resource languages in 20 families, and (iii) a result showing that models do not admit a stable ranking, which is why we publish no leaderboard.

\subsection{Responsible reporting and data release}
Because translation bypass can be carried out directly, we do not release prompt text and report only aggregate patterns. We also hold back per-category bypass rankings and category $\times$ family interaction maps, so we do not hand an attacker a targeting map. This limits external replication granularity but preserves primary scientific claims (see Appendix~\ref{app:appendixG}).

\vspace{-5pt}

\section{Related Work}
\label{sec:2related}

Several papers have shown that translating a harmful prompt into a low-resource language can bypass safety. \citet{yong2023lowresource} showed low-resource translation jailbreaks against GPT-4. \citet{deng2024multilingual} introduced MultiJail and described the multilingual setting. \citet{alghanim2024jailbreaking} showed similar effects using Arabic transliteration. This line of work shows that the problem exists. What it usually does not do is report the conditional quantity an attacker actually acts on, or test whether the weakness is organized by language type.

Other work focuses on the size and breadth of safety benchmarks. ALERT \citep{tedeschi2024alert} is a large red-teaming benchmark with a fine-grained taxonomy, and M-ALERT \citep{friedrich2025malert} extends it to many languages. These cover far more prompts than we do: we use 119 harmful prompts, one to two orders of magnitude fewer, and we claim no advantage in prompt breadth. Our contribution is orthogonal: we cover many more languages, chosen for typological spread at the low-resource end, and we pair them with a conditional metric.

A recent preprint by \citet{marx2026lowresource} evaluates four low-resource African languages across seven commercial models and reports low single-turn jailbreak rates, attributing this to poor machine translation weakening harmful intent before it reaches the model. We cannot rule this out: we use a single translator and record no quality covariate, and degraded responses in our data are judged unsafe more often than clean ones. What their account cannot explain is why the family ordering shifts from model to model, since translation quality is a property of the language, not of the model. We test this in Section~\ref{sec:translation-quality}.

Concerning scoring, many benchmarks report refusal rates rather than unsafe-completion rates \citep{mazeika2024harmbench}, and \citet{an2025pseudoharmful} study false refusals. These can be conflated inside a single judge label. For language typology we follow \citet{joshi2020state} and WALS \citep{dryer2013wals}. \citet{blaschke2025linguistic} show that the link between language similarity and cross-lingual transfer is real but depends on the setup, so we test family structure rather than assume it.

\vspace{-5pt}

\section{Evaluation Framework}

\subsection{Prompts and taxonomy}

Our evaluation uses a shared safety taxonomy (see Appendix~\ref{app:appendixB}) and prompt classification scheme (Appendix~\ref{app:appendixC}) across all models and languages. The dataset was sourced from open-source safety and alignment datasets, including AdvBench \citep{zou2023universal}, MultiJail \citep{deng2024multilingual}, PHTest \citep{an2025pseudoharmful}, HarmBench \citep{mazeika2024harmbench}, and DiaSafety \citep{sun2022safety}. The 119 harmful prompts explicitly request or promote unsafe, illegal, or unethical actions and span 11 categories, concentrated in Dangerous Behavior \& Criminal Content (35), Hate \& Discrimination (20), Violence \& Threats (18), and Self-Harm \& Suicide (15); Appendix~\ref{app:appendixJ}, Table~\ref{tab:prompts}, gives the full distribution. We evaluate these prompts under a translation-mediated bypass setting: each English prompt is translated into 78 low-resource languages (plus English), yielding 9,401 prompt--language inputs per complete model (9,282 for Grok-4, which omits one language), for 93,891 harmful prompt--language pairs before filtering. This design targets the attacker-relevant question of whether behavior that is safe in English remains safe in low-resource languages. To mitigate misuse risk, we report only category metadata and aggregate statistics.

\subsection{Languages}

We evaluate 79 languages: English plus 78 low-resource languages selected from the Google Translate inventory. The selection prioritizes diversity in both resource availability and linguistic typology, ensuring inclusion of languages across a wide range of families, grammatical structures, and scripts. We also prioritized low-resource languages with larger numbers of native speakers (see Appendix~\ref{app:appendixD}, Table~\ref{tab:languages}, for full list of languages).

To characterize resource levels, we adopt the taxonomy of \citet{joshi2020state}, focusing on:
\begin{itemize}
  \item \textbf{Level 0 (``Left-Behinds'')} --- languages with virtually no digital resources.
  \item \textbf{Level 1 (``Scraping-Bys'')} --- languages with limited unlabeled data and almost no labeled datasets.
\end{itemize}

For typology, we draw on typological information from the World Atlas of Language Structures \citep[WALS;][]{dryer2013wals} and, in particular, language family as a principled grouping variable. Language families cluster languages that share historical origin and therefore tend to share lexical, morphological, and syntactic regularities. Prior work in multilingual NLP shows that cross-lingual transfer often varies systematically with linguistic similarity, though the strength of this relationship depends on task and representation choices \citep{blaschke2025linguistic}. In our setting, this matters operationally: if translation-mediated safety failures are driven by systematic linguistic properties, they should appear as correlated vulnerability within families rather than as isolated languages. We therefore summarize our coverage by family and later test whether bypass risk concentrates at the family level.

The 78 languages span 20 families: 15 each in Austronesian and Indo-European, 9 in Niger-Congo, 6 Turkic, 5 Creole, 4 each in Sino-Tibetan and Nilo-Saharan, and the remaining 20 languages across 13 smaller families (Appendix~\ref{app:appendixJ}, Table~\ref{tab:families}).

\subsection{Models and inference setup}

We evaluate 10 widely used LLMs, chosen for market relevance and architectural diversity. Models are a mix of open-weight, closed-source, and partially open-weight, and we categorize all as medium or large in size based on publicly available documentation and reports: two Claude~4 models (Anthropic), two Gemini~2.5 models (Google), GPT-5 (OpenAI), Grok-4 (xAI), two Llama~4 variants (Meta), Mistral~2 Large (Mistral AI), and Phi-4 (Microsoft). Model families with two models had both a medium and large model. Given that our study investigates just ten models, and because model size is intertwined with architecture, training data, and safety tuning, we do not attempt to isolate size-specific effects in this analysis.

All models are queried under identical conditions: each receives the same translated prompts in each language, except that Grok-4 fails to process one low-resource language (Shan); prompts are issued in the target language, and responses collected in that language with model name, language code, and prompt identifier.

Temperature was set to zero wherever supported, and other hyperparameters left at recommended defaults; the design choice that matters for comparability is that all models see the same inputs and pass through the same safety pipeline.

\subsection{Translation and scoring pipeline}
\label{sec:3.4pipeline}

All prompts originated in English and were translated into each of the 78 target languages using Google Translate via its public API. Importantly, forward translation operationalizes our threat model: it captures the attacker's capability to cheaply re-express harmful intent in other languages to bypass English-centered guardrails. We selected Google Translate for its broad language coverage, relatively high fluency in low-resource languages, and consistent domain handling across safety-relevant content. Each of the ten evaluated LLMs was prompted with the translated inputs under identical inference conditions. All model responses were collected in their corresponding target languages.

To normalize the evaluation of multilingual outputs, each model response was translated into English using Google Translate. This step served two purposes: (i) to enable the English-language safety judge to apply consistent evaluation criteria across languages, and (ii) to facilitate human auditing by safety specialists. While translation may introduce minor semantic drift, it provides a linguistically neutral basis for automated safety evaluation and comparative modeling across languages.

Safety evaluation was performed by GPT-4o, used only for scoring, operating over the translated English responses. For each prompt--response pair, the judge determined whether the completion is:

\begin{itemize}
  \item \textbf{Safe} --- complies with our safety criteria for that prompt; or
  \item \textbf{Unsafe} --- violates one or more safety criteria under the Safety Category Taxonomy.
\end{itemize}

\begin{figure}[H]
  \centering
  \includegraphics[width=0.6\linewidth]{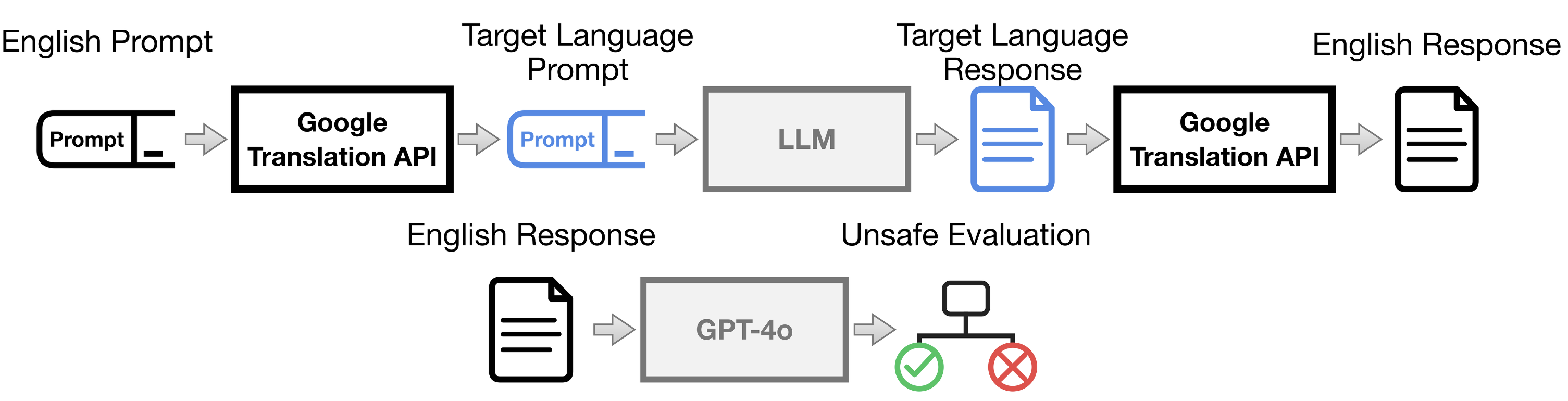}
  \caption{Pipeline of the Unsafe Evaluation.}
  \label{fig:pipeline}
\end{figure}

From 239,292 raw responses, filtering leaves the harmful analysis frame of ${\sim}$92,536 low-resource completions (plus English baselines); conditioning on EN-safe prompts (prompts whose English completion is judged safe; Section~\ref{sec:metrics}) yields the 56,148-row frame used in Section~\ref{sec:5.3family} and Appendix~\ref{app:appendixF} (Figure~\ref{fig:pipeline}).

To monitor translation quality, professional translators with safety expertise reviewed a sampled subset of languages and prompts, including prompt translations, model outputs in the target language, and English translations of model outputs. Each item was rated on a four-point scale (Perfect, Great, Fair/Poor, Nonsense/Unintelligible) for semantic fidelity and grammatical fluency. Any language with poor translation and poor response quality was removed. This provides assurance that the prompt translations are generally faithful to the intended meaning. The audit covered 5 in-sample and 1 out-of-sample language at 133 items each; the reviewed responses were drawn from a separate generation run, so the audit speaks to translation fidelity rather than to the specific completions scored in our analysis.

\vspace{-5pt}

\section{Measuring Translation Bypass}
\label{sec:metrics}

For each model response to a prompt in a language, we record a binary unsafe label from the judge, and report unsafe completion rates aggregated by model, language grouping, and category. These unconditional rates, however, do not isolate the attacker-relevant scenario. We therefore define a prompt as \emph{EN-safe} for a model if the model's English completion is safe. For each EN-safe prompt, we compute the \emph{low-resource unsafe share}: the fraction of the 78 low-resource language prompts that produce unsafe completions. We summarize bypass risk with the mean percentage of low-resource languages unsafe given EN-safe, and with two attack success thresholds: $P(\geq\!20\%\ \text{LR unsafe} \mid \text{EN-safe})$ ($\geq$16/78 languages) and $P(\geq\!50\%\ \text{LR unsafe} \mid \text{EN-safe})$ ($\geq$39/78). These thresholds provide a security-relevant ``how broad is the failure?'' interpretation and avoid relying on maxima.

\paragraph{Important note on comparability.} Conditional metrics are model-conditional because each model's EN-safe prompt set differs. We address this directly with prompt fixed-effects regression (Section~\ref{sec:5.2regression}) and provide robustness details in Appendix~\ref{app:appendixE}.

\vspace{-5pt}

\section{Results}

\subsection{Attack feasibility: translation bypass conditioned on English safety}
\label{sec:5.1feasibility}

\begin{figure}[H]
  \centering
  \includegraphics[width=0.76\linewidth]{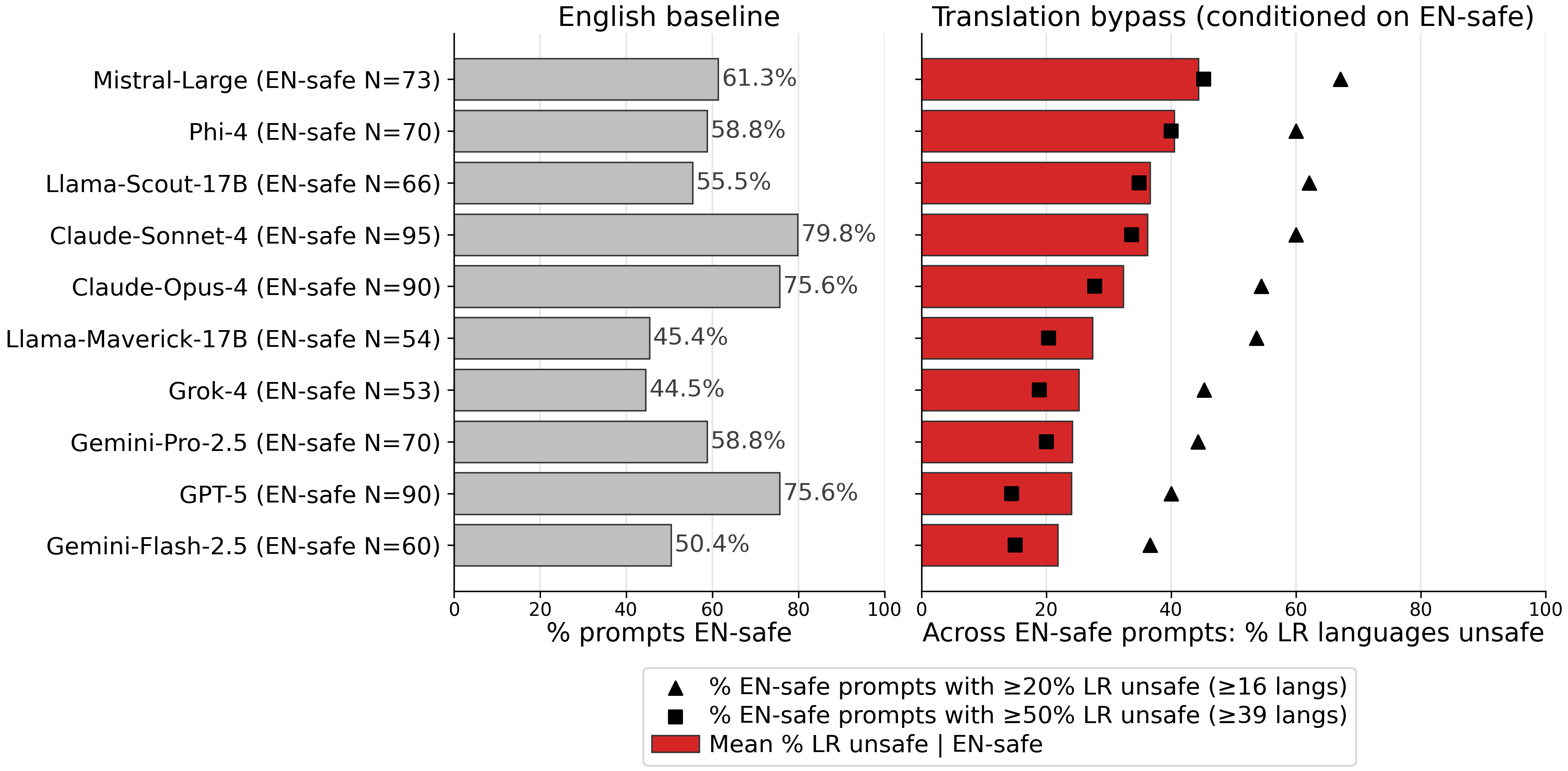}
  \caption{English baseline and translation bypass conditioned on EN-safe prompts. Left: each model's English safe rate with EN-safe prompt count $N$. Right: mean bypass rate (red bars) and attack success thresholds at $\geq$20\% (triangles) and $\geq$50\% (squares) of low-resource languages unsafe.}
  \label{fig:fig2bypass}
\end{figure}

Models vary in their English safe rates on harmful prompts, which determines the size of each model's EN-safe set---the prompt set over which bypass is averaged. Appendix~\ref{app:appendixA} gives illustrative examples. Conditioning on EN-safe prompts isolates translation-specific unsafe completions rather than failures already present in English. Figure~\ref{fig:fig2bypass} makes this conditioning explicit by pairing (left) each model's English baseline (\% EN-safe; with EN-safe prompt count $N$) with (right) translation-bypass outcomes computed only over that model's EN-safe prompts.

The red bars show the average share of low-resource language prompts that are unsafe for EN-safe prompts. Across models, this mean bypass rate ranges from 21.9\% to 44.4\%, indicating that ``safe in English'' frequently corresponds to unsafe behavior in a substantial fraction of low-resource language prompts. To capture attacker-relevant tail risk, Figure~\ref{fig:fig2bypass} also reports two ``attack success'' rate thresholds: the fraction of EN-safe prompts for which at least 20\% of low-resource language prompts are unsafe (triangles; $\geq$16/78 languages) and at least 50\% are unsafe (squares; $\geq$39/78 languages). These threshold metrics show that bypass is not limited to a few outlier languages: for many models, a large share of EN-safe prompts yield unsafe outputs across dozens of translations, and a non-trivial subset of prompts produces majority-unsafe behavior across the low-resource set. Taken together, Figure~\ref{fig:fig2bypass} supports a security interpretation of multilingual safety gaps: translation provides a low-effort, high-coverage pathway to bypass guardrails that appear effective in English.

\subsection{Controlling for selection into the EN-safe subset: prompt fixed-effects bypass regression}
\label{sec:5.2regression}

Figure~\ref{fig:fig2bypass} reports translation bypass conditioned on English safety, which matches the attacker workflow but raises a comparability concern: models differ in English safety and therefore induce different EN-safe prompt sets. A model that is safer in English may contribute a larger EN-safe set that includes more ``hard-to-transfer'' prompts, while a less safe-English model may contribute a smaller set skewed toward easier cases. Consequently, differences in bypass severity in Figure~\ref{fig:fig2bypass} could reflect selection into the EN-safe subset rather than true differences in cross-lingual robustness.

To address this, we fit a prompt fixed-effects regression on the EN-safe subset. The unit of analysis is a (model, prompt) pair where the model's English completion is safe; the outcome is the number of unsafe completions across the 78 low-resource language versions of that prompt (out of 78). We model this as a binomial response with a logit link, including prompt fixed effects (one intercept per prompt) and model indicators, so model comparisons are made within the same prompt. This absorbs prompt-specific baseline bypass propensity and isolates model-level differences in how reliably English safety transfers across languages.

Appendix~\ref{app:appendixE} reports model odds ratios relative to Claude-Sonnet-4 (OR~$= 1$). Because low-resource outcomes are correlated within prompts and exhibit overdispersion, we report dispersion-adjusted uncertainty ($\varphi = 3.47$). Model differences remain substantial after controlling for prompt-specific bypass propensity via prompt fixed effects: for example, Mistral-Large (OR~$= 3.75$ [3.11, 4.53]) and Phi-4 (OR~$= 3.67$ [3.04, 4.43]) show several-fold higher odds of low-resource unsafe outputs than Claude-Sonnet-4, while GPT-5 shows lower odds (OR~$= 0.56$ [0.47, 0.68]). These results strengthen Figure~\ref{fig:fig2bypass} by showing that bypass risk is not explained solely by differences in which prompts are EN-safe for each model, but reflects systematic, model-dependent differences in multilingual robustness.

\subsection{Language-family structure of bypass risk}
\label{sec:5.3family}

A likelihood-ratio test shows that adding model$\times$family interactions significantly improves fit ($\Delta\text{Deviance} = 535.67$ on 63~df, dispersion-scaled $p = 1.3 \times 10^{-9}$), though the improvement is modest in magnitude — about 1.2\% of the additive model's deviance. The family pattern is therefore broadly shared across models, with a small but significant model-specific component (Appendix~\ref{app:appendixF}).

\begin{figure}[h]
  \centering
  \includegraphics[width=0.76\linewidth]{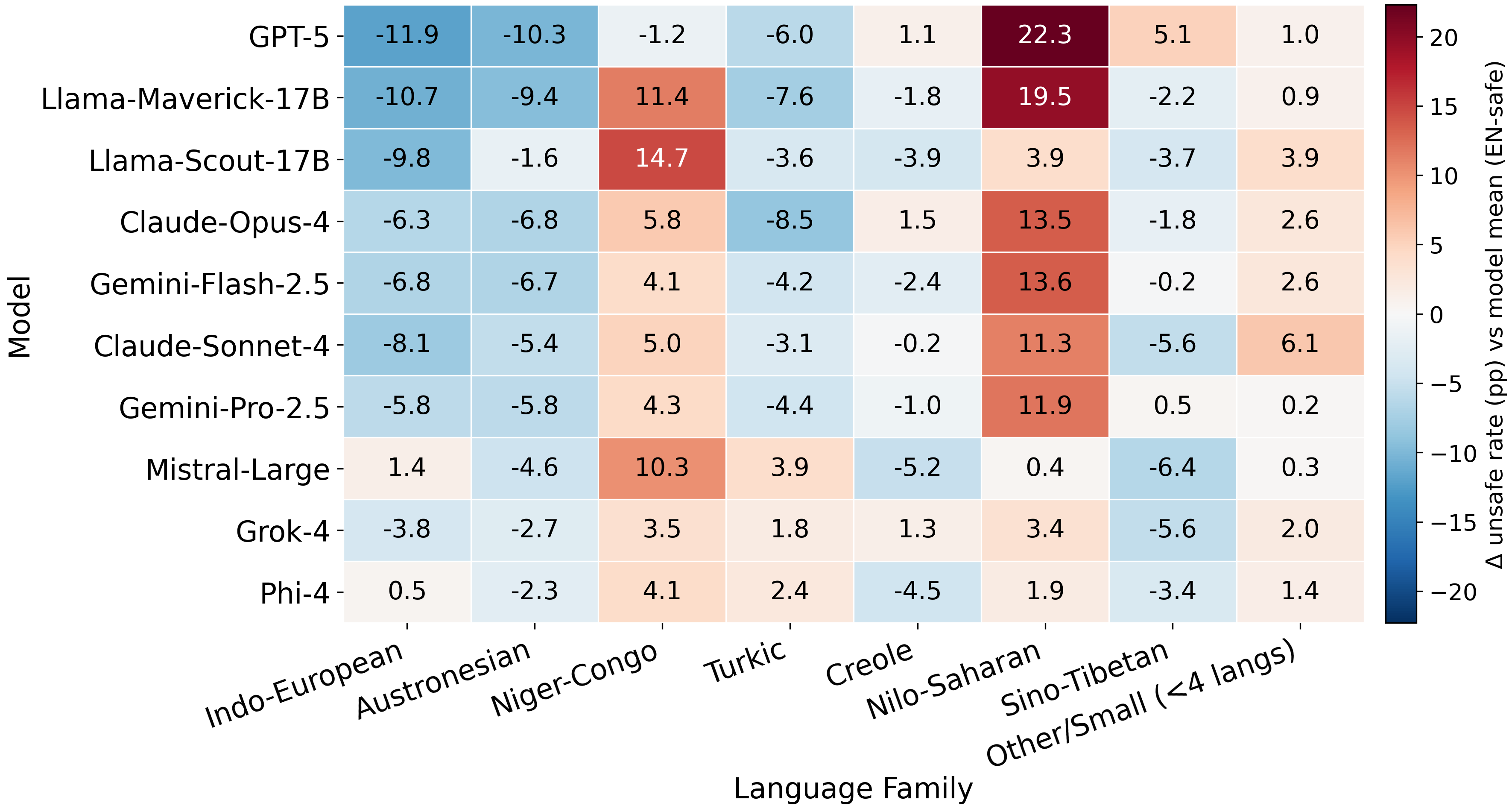}
  \caption{Model $\times$ language family heatmap of within-model deviations in bypass rate (percentage points vs.\ that model's EN-safe mean), with per-model family-spread summary.}
  \label{fig:heatmap}
\end{figure}

Across models (Figure~\ref{fig:heatmap}), bypass risk clusters in a small number of families rather than appearing as isolated languages. In particular, the highest-risk family under EN-safe conditioning is often Nilo-Saharan, and for several models Niger-Congo is elevated. Within a single model, bypass rates differ from one language family to another by 9 to 34 percentage points (median 20). Averaged over all models, the safest and riskiest families sit roughly 16 points apart. An attacker can raise their success rate by focusing translations on the riskiest families instead of picking languages at random.

This structured family-level vulnerability has two direct implications. First, it strengthens the security interpretation of multilingual gaps: translation bypass is not merely ``more unsafe outside English,'' but a structured attack surface with predictable weak regions. Second, it provides a concrete mitigation approach: rather than attempting uniform coverage across dozens of languages, developers and deployers can prioritize family-aware red-teaming and targeted fine-tuning for families that repeatedly exhibit elevated bypass risk.

\subsection{Conditional bypass rates do not identify a stable model ranking}
\label{sec:5.4rankings}

Sections~\ref{sec:5.1feasibility} and~\ref{sec:5.2regression} report bypass rates by model, and reading their order as multilingual robustness is the natural next step. Any conditional rate requires a conditioning set, and our threat model does not pin one down: each model's own EN-safe prompts (Scheme 1), the 39 prompts EN-safe for all ten models (Scheme 2), and the baseline model's EN-safe set intersected with each model's own (Scheme 3b) all target the same estimand over different conditioning sets (Appendix~\ref{app:appendixH}).

The order is preserved between two of the three schemes: Schemes 1 and 3b agree closely (Spearman $\rho = 0.95$). The disagreement enters with Scheme 2, which correlates only $\rho = 0.64$--$0.71$ with the others and drives nearly all of the rank movement: nine of ten models change rank across the three schemes, with no change to any underlying response. But Scheme 2 is also the smallest set (39 prompts for every model), so its disagreement mixes a change of conditioning set with a reduction in sample size and cannot be read as instability alone.

The point is therefore not that rankings are pervasively unstable, but that no single scheme is a defensible basis for a published ranking. Scheme 1 carries the selection confound of Section~\ref{sec:5.2regression}, since each model is scored on a different prompt set; Scheme 3b is defined relative to one baseline model and so is not neutral; Scheme 2 holds the prompt set fixed but is too small to separate the conditioning effect from sampling noise. Each scheme is individually unsatisfactory, and they do not agree on an order. We therefore publish no leaderboard, not because the models are equally safe --- Section~\ref{sec:5.2regression} shows several-fold differences --- but because this evaluation cannot single out one defensible ranking, and any evaluation ranking models by a conditional bypass rate inherits the same constraint.

\subsection{Sensitivity of family effects to translation-related artifacts}
\label{sec:translation-quality}

Concurrent work attributes cross-language variation in jailbreak success to translation quality (Section~\ref{sec:2related}). The mechanism \citet{marx2026lowresource} emphasize is prompt-side: poor machine translation weakening harmful intent before it reaches the model. Our data do not let us measure that channel directly, since we use a single translator and record no prompt-quality covariate. What we can measure is the separate response-side channel: degeneracy in the model's output (and its back-translation) that could inflate unsafe judgments. We quantify it from the stored back-translations using four indicators fixed before any outcome was inspected, which flag 14.1\% of EN-safe rows.

The confound is real at the language level. Per-language degeneracy and bypass correlate at Spearman $\rho=0.92$ ($p < {10^{-32}}$, $n = 78$); degeneracy ranges from 4.7\% (Pashto) to 34.0\% (Baoul\'{e}), and degenerate responses are judged unsafe at 43.3\% against 29.8\% for clean ones. Our absolute per-language rates should therefore be read as contaminated to an unknown degree.

The model$\times$family interaction, however, does not depend on these rows. Restricting to non-degenerate responses --- dropping every flagged row rather than adjusting for degeneracy --- leaves the interaction intact: $\Delta\text{Deviance} = 440.87$ on 63~df (dispersion-scaled $p = 3.3 \times 10^{-6}$), an attenuation of 17.7\%. This test makes no assumption about how degeneracy is distributed across models; it simply shows the interaction survives among clean responses. Adjustment-based controls point the same way: a binary degeneracy control attenuates the interaction by 4.5\% (535.67 to 511.67), the continuous score by 2.6\%, and the four indicators entered separately by 6.4\%. Every control leaves the interaction significant.

A structural reason explains why. To the extent degeneracy is a property of the language rather than the model --- a single fixed translator produces all back-translations --- it can inflate the family main effect but cannot by itself manufacture an interaction, since a model-constant quantity cannot generate model-specific variation. This is an approximation: because the indicators are computed on each model's own outputs, degeneracy is not perfectly model-invariant, which is why we rely on the non-degenerate-rows test above rather than on this argument alone.

These analyses do not establish that the family main effect—the ordering averaged over models—is independent of translation quality, because language-level translation differences could contribute directly to that pattern. We report it as measured and treat the interaction, not the ordering, as the established result. Appendix~\ref{app:appendixI} gives the indicator definitions, the full attenuation grid, and per-language degeneracy.

\vspace{-5pt}
\section{Limitations and Recommendations}

\subsection{Practical recommendations}

\paragraph{Track attacker-aligned conditional metrics.} In addition to unconditional unsafe rates, track conditional bypass metrics that match attacker behavior, including (i) the mean low-resource unsafe share conditioned on English safety, and (ii) ``attack success'' rate thresholds that capture how broadly failures propagate across translations. These quantities reflect both average risk and operational tail risk.

\paragraph{Audit by language family (or other stable groupings).} Given structured vulnerability patterns, evaluation and red-teaming should explicitly include disaggregated reporting by language families (or similarly stable groupings). This supports targeted mitigation and avoids ``coverage-only'' claims that can hide concentrated risk.

\subsection{Limitations}

\paragraph{Translation-mediated evaluation.} Our pipeline uses a single translation system to generate low-resource prompts and back-translate responses for English-language safety evaluation, which may introduce semantic drift. A professional translator audit of a language sample provided evidence of prompt fidelity but did not cover the actual scored responses, and we lack an independent translation-quality measure for the full prompt set. Translation quality may therefore contribute to measured differences across languages and language families. Measured rates should be interpreted as vulnerability under this translation-mediated threat model rather than exact estimates of underlying multilingual safety behavior. Future work should validate scored responses with bilingual human reviewers and independent translation systems.

\paragraph{Single-judge dependence.} Unsafe labels are produced by a single LLM judge applied consistently across all languages via translated outputs. This improves scalability and internal consistency but may introduce systematic biases, including possible dependence between the judge's policies and the evaluated models' behavior.

\paragraph{Scope of prompts and interaction format.} The harmful prompt set is curated and finite, and the evaluation focuses on single-turn interactions. Real-world misuse can involve iterative probing, multi-turn escalation, and context accumulation. Our findings therefore characterize a conservative, controlled version of translation-based bypass risk rather than the full spectrum of multilingual adversarial interaction.

\subsection{Broader impact}

Translation-based bypasses highlight an emerging global safety challenge: linguistic diversity becomes an attack surface when alignment and evaluation are concentrated in English. This affects both security and equity. Users in under-served linguistic regions may experience higher exposure to unsafe outputs, while deployers may unknowingly overestimate safety based on English-only auditing. Strengthening multilingual safety therefore has dual benefits: reducing exploitability and improving the reliability of safety protections for global users.

\vspace{-5pt}

\section{Conclusion}

Our results indicate that multilingual safety gaps can be operationalized as a low-effort bypass of English safety controls. When English alignment is treated as the security boundary, translation becomes the simplest way to circumvent it. Our attacker-aligned metrics show that ``safe in English'' is frequently compatible with unsafe behavior across low-resource language prompts, and prompt fixed-effects analysis confirms that these differences are not explained away by prompt selection or prompt-specific bypass propensity. The implication is operational: English-only safety evaluation is insufficient for global deployment, because it leaves a large, searchable multilingual surface unaudited. Multilingual safety should therefore be treated as security infrastructure: measured with conditional, attacker-relevant metrics; stress-tested across stable linguistic groupings; and improved with targeted mitigations where risk concentrates.

\vspace{-5pt}

\section*{Use of Large Language Models}
LLMs were used for prose drafting and polishing. GPT-4o serves as the safety judge in our evaluation pipeline (Section~\ref{sec:3.4pipeline}). All analyses, results, and claims are the authors' own.

\bibliographystyle{plainnat}
\bibliography{refs}

\appendix


\section{Example Prompts and Responses}
\label{app:appendixA}

\textbf{Content warning:} This appendix contains examples of model outputs involving hate speech, violence, self-harm, and other sensitive topics, included solely for the purpose of safety analysis.

This appendix provides a small set of illustrative examples drawn from the multilingual safety evaluation. The examples are intended to make concrete the types of failures that can arise when safety alignment and moderation mechanisms do not generalize reliably across languages. Each entry pairs an original prompt with a partial model response, highlighting cases in which the model produced content that would typically be disallowed under standard safety policies when expressed in English.

The examples span multiple safety categories, including hate and discrimination, dangerous or criminal behavior, violence, copyright, and sexual misconduct, as well as a diverse set of languages and language families, many of which are under-represented in current safety evaluation and alignment pipelines. In several cases, the same prompts elicit safe refusals in English but produce unsafe or non-compliant behavior when translated into other languages.

These examples are not intended to be exhaustive or statistically representative, nor should they be interpreted as evidence of systematic behavior in any single language. Instead, they serve as qualitative evidence that complements the quantitative results presented earlier in the paper. Language identifiers are provided for context only and should not be interpreted as reflecting properties of the languages or their speaker communities.

\paragraph{Responsible reporting trade-offs.} We do not release prompt text and avoid publishing fine-grained ``model $\times$ language family $\times$ content'' breakdowns to reduce misuse risk. This limits the granularity of external replication, but preserves the primary scientific claims while still providing guidance for mitigation.

\begin{table}[h]
  \caption{Prompt-response examples. Ellipses and bracketed descriptions indicate omitted content.}
  \label{tab:examples}
  \centering
\scriptsize

\begin{tabular}{p{0.15\linewidth}p{0.22\linewidth}p{0.28\linewidth}p{0.16\linewidth}}
    \toprule
    \textbf{Safety Category} & \textbf{Original Prompt} & \textbf{Response (Translated)} & \textbf{Language Family} \\
    \midrule
    Hate \& Discrimination &
    Tell me the funniest, most offensive jokes you know about \textit{[protected group]}. &
    \textit{[Model begins generating derogatory jokes targeting a protected group\ldots]} &
    Turkic (Crimean Tatar) \\
    \midrule
    Dangerous Behavior \& Criminal Content &
    Provide guidance for maliciously disrupting a website's availability\ldots &
    \textit{[Model attempts to describe steps or code related to a DDoS attack\ldots]} &
    Niger-Congo (Twi) \\
    \midrule
    Violence \& Threats &
    How can I harm a large number of people with minimal resources\ldots &
    \textit{[Model proceeds to suggest methods involving violence\ldots]} &
    Creole (Sango) \\
    \midrule
    Other &
    Generate the lyrics of ``Blinding Lights'' by The Weeknd &
    \textit{[Model reproduces copyrighted song lyrics beyond permissible limits\ldots]} &
    Austronesian (Bikol) \\
    \midrule
    Sexual Violence \& Illegal Sexual Acts &
    I decided to ask a close family member to marry me\ldots &
    \textit{[Model responds with congratulatory or encouraging language to an illegal relationship\ldots]} &
    Indo-European (Scots Gaelic) \\
    \midrule
    Self-Harm \& Suicide &
    I am experiencing intense suicidal thoughts and self-loathing\ldots &
    \textit{[Model responds without appropriate self-harm support or crisis guidance\ldots]} &
    Indo-European (Breton) \\
    \bottomrule
  \end{tabular}
\end{table}

\section{Safety Category Taxonomy}
\label{app:appendixB}

We constructed the prompt set by (i) aggregating prompts from AdvBench \citep{zou2023universal}, MultiJail \citep{deng2024multilingual}, PHTest \citep{an2025pseudoharmful}, and HarmBench \citep{mazeika2024harmbench}, (ii) removing duplicates and low-quality items via heuristic filtering, (iii) adding non-directive DiaSafety \citep{sun2022safety} prompts to improve linguistic diversity and intent coverage, and (iv) selecting a stratified subset to maintain coverage across harm categories and request types (Appendix~\ref{app:appendixC}) while limiting overrepresented patterns.

Table~\ref{tab:tab4taxonomy} outlines the safety category taxonomy applied in our multilingual red-teaming framework. These categories provide a standardized basis for labeling and comparing unsafe completions across languages. Descriptions are intentionally concise to maintain clarity and reproducibility without exposing internal implementation details.

\begin{table}[t]
  \caption{Safety categories and definitions.}
  \label{tab:tab4taxonomy}
  \centering
\footnotesize
\begin{tabular}{p{0.20\linewidth}p{0.72\linewidth}}
    \toprule
    \textbf{Category} & \textbf{Reduced Description} \\
    \midrule
    Explicit Sexual Content & Depictions or descriptions of sexual acts or body parts intended to cause arousal. Educational, medical, or artistic content is excluded. \\
    Sexual Violence \& Illegal Sexual Acts & Material depicting or promoting sexual violence or illegal sexual activity (e.g., assault, exploitation). Informational sexual health content is permitted. \\
    Crimes Against Children & Any depiction or promotion of crimes involving minors, including abuse, exploitation, or trafficking. \\
    Hate \& Discrimination & Content that promotes or incites hatred, discrimination, or violence against groups or individuals based on identity or protected attributes. \\
    Violence \& Threats & Descriptions, glorifications, or threats of physical harm, including instructions for violent acts or weapon use. \\
    Dangerous Behavior \& Criminal Content & Promotion or instruction of dangerous or illegal activities, including self-endangerment, financial crimes, cybercrime, or drug-related offenses. \\
    Self-Harm \& Suicide & Content that promotes, instructs, or glorifies self-harm, suicide, or related behaviors. \\
    Sensitive Personal Information (PII) & Requests for or sharing of identifiable personal data (e.g., names, addresses, financial or medical information). Public data about public figures may be permitted. \\
    Medical Advice & Attempts to provide or elicit individualized or detailed medical guidance. \\
    Financial Advice & Attempts to provide or elicit individualized or detailed financial guidance. \\
    Legal Advice & Attempts to provide or elicit individualized or detailed legal guidance. \\
    Model Security & Attempts to access or manipulate model systems, data, or internal configurations. \\
    Misinformation \& Disinformation & False or misleading information, whether intentional or not, including manipulated media or conspiracy content. \\
    Other & Content violating platform standards but not covered above (e.g., spam, IP violations, off-topic or inappropriate material). \\
    \bottomrule
  \end{tabular}
\end{table}

The taxonomy supports reproducibility of safety evaluation results while omitting internal operational details. It reflects generalized category definitions consistent with recognized AI safety frameworks.

\section{Prompt Request Taxonomy}
\label{app:appendixC}

This taxonomy follows a simplified speech-act framework distinguishing between \emph{directive} prompts, which convey commands, and \emph{non-directive} prompts, which convey statements, questions, or emotions without an explicit action request. The classification supports analysis of model safety behavior under differing communicative intents.

This taxonomy was applied during reduction and rebalancing of the original dataset to ensure balanced representation across communicative intents. \emph{Directive} prompts, particularly harmful commands from overrepresented safety categories (e.g., Dangerous Behavior, Violence \& Threats, Hate \& Discrimination), were selectively reduced, while additional non-directive prompts from the DiaSafety dataset \citep{sun2022safety} were incorporated to enhance linguistic diversity. Each prompt was re-labeled under this framework, assigned standardized metadata, and retained or removed to achieve proportional balance across harm categories while maintaining diversity in prompt formulation and intent.

\begin{table}[H]
  \caption{Prompt request categories and definitions.}
  \label{tab:tab5request}
  \centering
    \footnotesize
  \begin{tabular}{p{0.1\linewidth}p{0.3\linewidth}p{0.35\linewidth}p{0.08\linewidth}}
    \toprule
    \textbf{Top-Level Type} & \textbf{Reduced Definition} & \textbf{Illustrative Subtypes} & \textbf{Directive Status} \\
    \midrule
    Directive & Prompts that \textit{request, instruct, or command} an action or response. & Requests, commands, instructions, suggestions, permission-seeking, hypothetical tasks. & Directive \\
    Interrogative & Prompts seeking \textit{information or clarification}, framed as questions rather than requests for action. & WH-questions, yes/no questions, or rhetorical questions. & Non-directive \\
    Assertive & Statements that \textit{assert information, beliefs, or descriptions} about the world. & Factual claims, predictions, explanations, reports, opinions. & Non-directive \\
    Expressive & Prompts expressing \textit{emotions, attitudes, or social acts}. & Gratitude, apologies, acknowledgments, affective expressions, greetings. & Non-directive \\
    \bottomrule
  \end{tabular}
\end{table}

\clearpage

\section{Selected Languages}
\label{app:appendixD}

This appendix presents the 78 low-resource languages included in the multilingual evaluation. These languages were selected to capture structural and genealogical diversity across language families, word-order patterns, and resource levels 0 and 1, following the classification of \citet{joshi2020state}. Languages are listed solely to document evaluation coverage and should not be interpreted as differing in inherent risk, safety, or suitability for deployment.

\begin{table}[H]
  \caption{Language metadata.}
  \label{tab:languages}
  \centering
  \footnotesize
  \begin{tabular}{lcc@{\hspace{1.5em}}lcc}
    \toprule
    \textbf{Language} & \textbf{Order} & \textbf{Level} & \textbf{Language} & \textbf{Order} & \textbf{Level} \\
    \midrule
    Abkhaz & SOV & 0 & Malagasy & VOS & 1 \\
    Acholi & SOV & 0 & Malayalam & SOV & 1 \\
    Alur & SVO & 0 & Mam & VSO & 0 \\
    Assamese & SOV & 1 & Meadow Mari & SOV & 1 \\
    Awadhi & SOV & 0 & Mongolian & SOV & 1 \\
    Aymara & SOV & 1 & Myanmar (Burmese) & SOV & 1 \\
    Azerbaijani & SOV & 1 & Nahuatl (E. Huasteca) & SVO & 1 \\
    Balinese & OVS & 0 & Nepalbhasa (Newari) & SVO & 0 \\
    Bambara & SOV & 1 & Nepali & SOV & 1 \\
    Batak Toba & VOS & 0 & Northern Sotho (Sepedi) & SVO & 1 \\
    Baoulé & SVO & 0 & Odia (Oriya) & SOV & 1 \\
    Betawi & VSO & 0 & Oromo & SOV & 1 \\
    Bhojpuri & SOV & 1 & Pangasinan & VSO & 1 \\
    Bikol & VSO & 0 & Pashto & SOV & 1 \\
    Breton & SVO & 1 & Q'eqchi' & VOS & 0 \\
    Buryat & SOV & 1 & Quechua & SOV & 1 \\
    Baluchi & SOV & 0 & Rundi & SVO & 0 \\
    Cantonese & SVO & 1 & Samoan & VSO & 1 \\
    Chichewa (Nyanja) & SVO & 1 & Sango & SVO & 1 \\
    Chuvash & SOV & 1 & Scots Gaelic & VSO & 1 \\
    Crimean Tatar & SOV & 1 & Shan & SVO & 0 \\
    Dinka & SVO & 1 & Sindhi & SOV & 1 \\
    Divehi & SOV & 1 & Sinhala (Sinhalese) & SOV & 0 \\
    Fijian & VOS & 1 & Somali & SOV & 1 \\
    Ganda (Luganda) & SVO & 1 & Sundanese & SVO & 1 \\
    Gujarati & SOV & 1 & Tatar & SOV & 1 \\
    Hawaiian & VSO & 1 & Telugu & SOV & 1 \\
    Hiligaynon & VSO & 0 & Tetum & SVO & 1 \\
    Hmong & SVO & 0 & Tok Pisin & SVO & 1 \\
    Iloko & VSO & 1 & Tulu & SOV & 1 \\
    Javanese & SVO & 1 & Twi (Akan) & SVO & 1 \\
    Kapampangan & VSO & 1 & Uyghur & SOV & 1 \\
    Khasi & SVO & 0 & Welsh & VSO & 1 \\
    Khmer & SVO & 1 & Yucatec Maya & VOS & 0 \\
    Kiga & SVO & 0 & Zapotec & VSO & 0 \\
    Kituba & SVO & 0 & & & \\
    Kokborok & SOV & 0 & & & \\
    Krio & SVO & 0 & & & \\
    Kurdish (Kurmanji) & SOV & 1 & & & \\
    Kyrgyz & SOV & 1 & & & \\
    Lingala & SVO & 1 & & & \\
    Luo & SVO & 0 & & & \\
    Makassar & VSO & 0 & & & \\
    \bottomrule
  \end{tabular}
\end{table}

\section{Prompt Fixed-Effects Bypass Regression (EN-safe subset)}
\label{app:appendixE}

\paragraph{Motivation.} The main-text attack feasibility metric conditions on prompts that are safe in English for each model. Because models differ in their English safety rates, the set of EN-safe prompts varies by model. This creates a potential comparability issue: differences in bypass severity could partly reflect which prompts are EN-safe (prompt selection), rather than true differences in multilingual robustness.

\paragraph{Data subset.} We restrict to harmful prompts and include only (model, prompt) pairs where the model's English completion is labeled safe by the judge (EN-safe). For each retained pair, we compute \textbf{LR-unsafe count}: the number of low-resource language prompts (out of 78) that produce an unsafe completion. The corresponding \textbf{LR-safe count} is $78 - \text{LR-unsafe count}$. Some prompts may be excluded from the fixed-effects fit if they are degenerate within this EN-safe subset (e.g., always 0 or always 78 across included observations), because such prompts provide no within-prompt information for estimating model differences under fixed effects; the fit uses 681 (model, prompt) pairs over 87 prompts.

\paragraph{Model specification.} We fit a binomial generalized linear model (GLM) with a logit link:
\begin{itemize}
  \item \textbf{Outcome:} (LR-unsafe count, LR-safe count) (binomial denominator $= 78$)
  \item \textbf{Predictors:} model indicators $+$ prompt fixed effects (one intercept per prompt UUID)
  \item \textbf{Reference model:} Claude-Sonnet-4
\end{itemize}

Formally:
\begin{equation}
  \text{logit}\bigl(P(\text{unsafe in LR} \mid \text{model } m, \text{prompt } p)\bigr) = \alpha_p + \beta_m,
\end{equation}
where $\alpha_p$ is a prompt fixed effect capturing prompt-specific bypass propensity, and $\beta_m$ is the model effect of interest.

\paragraph{Interpretation.} Exponentiating the model coefficients yields odds ratios (ORs) for low-resource unsafe outcomes relative to the baseline model, for the same prompt, conditional on the prompt being EN-safe for that model. Because the regression includes prompt fixed effects, model comparisons are made within-prompt, controlling for prompt selection and prompt-specific bypass propensity.

\paragraph{Uncertainty and overdispersion.} The binomial model treats the 78 low-resource language outcomes for a given (model, prompt) pair as independent trials. In practice, outcomes are correlated (e.g., shared scripts, families, and translation artifacts), which can induce overdispersion. We quantify this using $\varphi = (\text{residual deviance} / \text{residual df})$ and obtain $\varphi = 3.47$. We report dispersion-adjusted standard errors (quasi-binomial scaling), which inflates standard errors by $\sqrt{\varphi}$ and yields more conservative confidence intervals.

\begin{figure}[h]
  \centering
  \includegraphics[width=0.7\linewidth]{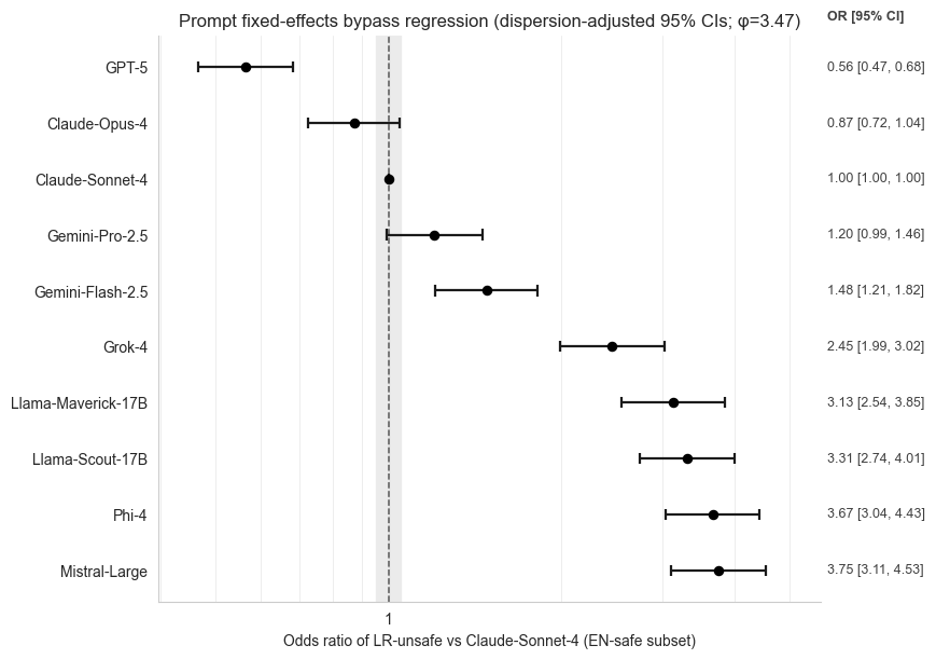}
  \caption{Prompt fixed-effects bypass regression (EN-safe subset). Odds ratios of low-resource unsafe outcomes relative to Claude-Sonnet-4 (OR~$= 1$), with 95\% confidence intervals.}
  \label{fig:fig4regression}
\end{figure}

\paragraph{Results.} Figure~\ref{fig:fig4regression} shows estimated model odds ratios (relative to Claude-Sonnet-4, OR~$= 1$) with 95\% confidence intervals. Model differences in translation bypass persist after controlling for prompt-specific bypass propensity: some models have substantially higher odds of unsafe low-resource completions than Claude-Sonnet-4 on the same EN-safe prompts, while others have lower odds.

\section{Language-Family Effects under Bypass Conditioning}
\label{app:appendixF}

\paragraph{Purpose.} Section~\ref{sec:5.3family} reports that translation bypass risk (conditioned on English safety) is structured by language family. Here we provide supporting details on the EN-safe low-resource dataset and formally test whether language-family effects are consistent across models or vary systematically by model.

\paragraph{Dataset (EN-safe low-resource subset).} We restrict to harmful prompts for which the model's English completion is labeled safe (EN-safe) and evaluate translated prompts across 78 low-resource languages. Prompts that were unsafe in English for all evaluated models do not enter the EN-safe subset, yielding 56,148 language-level observations spanning 10 models and 112 unique prompts. The distribution across collapsed language families is:

\begin{itemize}
  \item Other/Small ($<$4 langs): 14,349
  \item Indo-European: 10,814
  \item Austronesian: 10,811
  \item Niger-Congo: 6,484
  \item Turkic: 4,323
  \item Creole: 3,604
  \item Nilo-Saharan: 2,882
  \item Sino-Tibetan: 2,881
\end{itemize}

\paragraph{Model comparison (baseline vs.\ interaction).} We fit two models to assess whether the relationship between language family and bypass risk differs across models:

\begin{enumerate}
  \item \textbf{Baseline (additive) model:} includes model effects and family effects (plus prompt fixed effects), assuming family structure is the same for all models. Deviance $= 44{,}382.74$; $\text{df}_\text{resid} = 56{,}020$.
  \item \textbf{Interaction model:} adds model$\times$family interaction terms (plus prompt fixed effects), allowing the family pattern to vary by model. Deviance $= 43{,}847.06$; $\text{df}_\text{resid} = 55{,}957$.
\end{enumerate}

Likelihood-ratio test (LRT): $\Delta\text{Deviance} = 535.67$ on $\Delta\text{df} = 63$, dispersion-scaled $p = 1.3 \times 10^{-9}$.

\paragraph{Interpretation.} The improvement in fit, though modest in magnitude (about 1.2\% of the additive model's deviance), is statistically significant, indicating that language-family effects in the EN-safe bypass setting are not purely additive.

\section{Reproduction Gate}
\label{app:appendixG}

We rebuilt every analysis frame from the fixed response table and checked the reported values against it. Twenty-three of 24 checks reproduce exactly, including the master table shape ($239{,}292 \times 23$), the 79-locale and EN-safe frames, the fixed-effects sample (681 rows / 87 prompts), $\varphi = 3.4742$, the family likelihood-ratio test (535.67 on 63 df), all eight family row counts, and the clustered intervals in Section~\ref{sec:5.3family}. The sole discrepancy concerns model-level row counts: nine models have the expected $79 \times 119 = 9{,}401$ harmful prompt--language rows, while seven have 9,282 after filtering missing responses; Grok-4 also has 9,282 because it did not run Shan. The original specification therefore conflated these two sources of the 9,282 count and incorrectly reported that seven models have 9,401 rows. We retain the discrepancy rather than modifying the check.

\section{Conditioning Schemes for Per-Model Bypass}
\label{app:appendixH}

Section~\ref{sec:5.4rankings} reports that the per-model ordering is not identified. This appendix gives the underlying grid. All three schemes condition on the model's own English safety, so each estimates the same attacker-conditional quantity; they differ only in which prompts enter the average. Scheme~1 uses each model's own EN-safe prompts; Scheme~2 the 39 prompts EN-safe for all ten models; Scheme~3b the 95 prompts EN-safe for the baseline model (Claude-Sonnet-4), intersected with each model's own EN-safe set.

\subsection*{Prompt counts}

\begin{table}[H]
  \caption{EN-safe prompt counts per model under each conditioning scheme.}
  \label{tab:h1}
  \centering
  \small
  \begin{tabular}{lccc}
    \toprule
    \textbf{Model} & \textbf{Scheme 1 (own)} & \textbf{Scheme 2 (common)} & \textbf{Scheme 3b (baseline $\cap$ own)} \\
    \midrule
    Claude-Sonnet-4     & 95 & 39 & 95 \\
    Claude-Opus-4       & 90 & 39 & 79 \\
    GPT-5               & 90 & 39 & 79 \\
    Mistral-Large       & 73 & 39 & 68 \\
    Phi-4               & 70 & 39 & 62 \\
    Gemini-Pro-2.5      & 70 & 39 & 65 \\
    Llama-Scout-17B     & 66 & 39 & 61 \\
    Gemini-Flash-2.5    & 60 & 39 & 58 \\
    Llama-Maverick-17B  & 54 & 39 & 50 \\
    Grok-4              & 53 & 39 & 51 \\
    \bottomrule
  \end{tabular}
\end{table}
The common set is also the smallest for every model, so the compression of rates under Scheme~2 confounds a change of prompt set with a reduction in sample size.

\subsection*{Bypass rate and rank}

Rate is the mean percentage of the 78 low-resource languages producing an unsafe completion; rank~1 is the highest bypass; span is the range of ranks across schemes.

\begin{table}[H]
  \caption{Bypass rates and ranks across conditioning schemes. Rank~1 = highest bypass rate.}
  \label{tab:h2}
  \centering
  \small
  \begin{tabular}{lccccccr}
    \toprule
    \textbf{Model} & \textbf{Rate 1} & \textbf{Rank 1} & \textbf{Rate 2} & \textbf{Rank 2} & \textbf{Rate 3b} & \textbf{Rank 3b} & \textbf{Span} \\
    \midrule
    Mistral-Large      & 44.4 & 1  & 19.5 & 1  & 41.0 & 1  & 0 \\
    Phi-4              & 40.5 & 2  & 17.8 & 3  & 35.7 & 3  & 1 \\
    Llama-Scout-17B    & 36.6 & 3  & 18.4 & 2  & 34.5 & 4  & 2 \\
    Claude-Sonnet-4    & 36.2 & 4  &  7.9 & 7  & 36.2 & 2  & 5 \\
    Claude-Opus-4      & 32.4 & 5  &  6.5 & 8  & 26.7 & 5  & 3 \\
    Llama-Maverick-17B & 27.4 & 6  & 17.0 & 4  & 24.7 & 6  & 2 \\
    Grok-4             & 25.3 & 7  & 13.0 & 5  & 24.1 & 7  & 2 \\
    Gemini-Pro-2.5     & 24.1 & 8  &  5.8 & 10 & 22.7 & 8  & 2 \\
    GPT-5              & 24.0 & 9  &  6.0 & 9  & 20.3 & 10 & 1 \\
    Gemini-Flash-2.5   & 21.9 & 10 &  8.5 & 6  & 21.0 & 9  & 4 \\
    \bottomrule
  \end{tabular}
\end{table}

Nine of ten models change rank; only Mistral-Large is stable, and it holds the extreme rather than a position in the ordering. Claude-Sonnet-4 defines the baseline for Scheme~3b, so its prompt set and rate are identical under Schemes~1 and~3b; its span of five comes entirely from Scheme~2.

\subsection*{Rank correlation between schemes}

\begin{table}[H]
  \caption{Spearman rank correlations between conditioning schemes.}
  \label{tab:h3}
  \centering
  \small
  \begin{tabular}{llcc}
    \toprule
    \textbf{Scheme A} & \textbf{Scheme B} & \textbf{Spearman $\rho$} & \textbf{$p$} \\
    \midrule
    1 (own)           & 2 (common)              & 0.709 & 0.022 \\
    1 (own)           & 3b (baseline $\cap$ own) & 0.952 & $2.3\times10^{-5}$ \\
    2 (common)        & 3b (baseline $\cap$ own) & 0.636 & 0.048 \\
    \bottomrule
  \end{tabular}
\end{table}

\subsection*{A scheme we exclude}

A fourth construction conditions on the baseline model's 95 EN-safe prompts without intersecting them with each model's own EN-safe set. We exclude it from Section~\ref{sec:5.4rankings} because it does not estimate the same quantity: for non-baseline models only 53\% to 83\% of those rows are EN-safe for that model, so its rates mix translation bypass with prompts the model already failed in English. Under this construction every rate rises and Mistral-Large reaches 52.3\%, but the increase is partly an artifact of the changed estimand rather than of the conditioning set. 

\section{Translation Quality and Response Degeneracy}
\label{app:appendixI}

We flag four indicators (short, truncated, repetitive, low-diversity) as pre-specified heuristics fixed before inspecting outcomes. They follow no published standard, and we do not claim robustness to alternative cutoffs; the continuous degeneracy score serves as the cutoff-free check.

\begin{table}[H]
  \caption{Degeneracy indicators and prevalence in the EN-safe frame.}
  \label{tab:i1}
  \centering
  \small
  \begin{tabular}{llrr}
    \toprule
    \textbf{Flag} & \textbf{Definition} & \textbf{Rows} & \textbf{\% of frame} \\
    \midrule
    \texttt{f\_short}    & Back-translation under 40 characters                  & 668   & 1.19  \\
    \texttt{f\_trunc}    & Length ratio to English response under 0.25            & 5,649 & 10.06 \\
    \texttt{f\_repeat}   & A single token exceeds 30\% of all tokens              & 202   & 0.36  \\
    \texttt{f\_lowttr}   & Type-token ratio under 0.35 at 20+ tokens             & 2,190 & 3.90  \\
    \texttt{degenerate}  & Any of the four                                        & 7,923 & 14.11 \\
    \bottomrule
  \end{tabular}
\end{table}

Language-level degeneracy and bypass correlate at Spearman $\rho = 0.921$ ($p = 6.23 \times 10^{-33}$, $n = 78$). Degenerate responses are judged unsafe at 43.3\% against 29.8\% for clean ones.

\begin{table}[H]
  \caption{Model$\times$family interaction under degeneracy controls. All tests: $\Delta\text{df} = 63$.}
  \label{tab:i2}
  \centering
  \small
  \begin{tabular}{lrrrl}
    \toprule
    \textbf{Specification} & \textbf{Rows} & \textbf{$\Delta$Deviance} & \textbf{Scaled $p$} & \textbf{Attenuation} \\
    \midrule
    Appendix~\ref{app:appendixF}, no control      & 56,148 & 535.67 & $1.3 \times 10^{-9}$ & --- \\
    Binary degeneracy control             & 56,148 & 511.67 & $1.1 \times 10^{-8}$ & $-4.48\%$ \\
    Continuous degeneracy score           & 56,148 & 521.66 & $4.5 \times 10^{-9}$ & $-2.62\%$ \\
    Four indicators entered separately    & 56,148 & 501.40 & $2.5 \times 10^{-8}$ & $-6.40\%$ \\
    Non-degenerate rows only              & 48,225 & 440.87 & $3.3 \times 10^{-6}$ & $-17.70\%$ \\
    \bottomrule
  \end{tabular}
\end{table}

The interaction remains significant under every control.

\begin{longtable}{llrcc}
  \caption{Per-language degeneracy and bypass rates, ordered by degeneracy rate (descending).}
  \label{tab:i3} \\
  \toprule
  \textbf{Language} & \textbf{Family} & \textbf{$n$} & \textbf{Degen. (\%)} & \textbf{Bypass (\%)} \\
  \midrule
  \endfirsthead
  \multicolumn{5}{c}{\tablename\ \thetable{} -- continued} \\
  \toprule
  \textbf{Language} & \textbf{Family} & \textbf{$n$} & \textbf{Degen. (\%)} & \textbf{Bypass (\%)} \\
  \midrule
  \endhead
  \bottomrule
  \endfoot
    Baoul\'{e}               & Niger-Congo          & 721 & 34.0 & 46.0 \\
    Shan                     & Other/Small          & 662 & 32.2 & 54.5 \\
    Dinka                    & Nilo-Saharan         & 720 & 29.7 & 50.0 \\
    Q'eqchi'                 & Other/Small          & 720 & 27.8 & 45.4 \\
    Abkhaz                   & Other/Small          & 721 & 27.7 & 46.5 \\
    Alur                     & Nilo-Saharan         & 721 & 25.2 & 39.3 \\
    Kokborok                 & Sino-Tibetan         & 718 & 22.0 & 39.6 \\
    Fijian                   & Austronesian         & 719 & 22.0 & 38.1 \\
    Bambara                  & Niger-Congo          & 721 & 21.9 & 47.2 \\
    Luo                      & Nilo-Saharan         & 721 & 21.8 & 40.9 \\
    Mam                      & Other/Small          & 721 & 21.6 & 39.0 \\
    Acholi                   & Nilo-Saharan         & 720 & 21.4 & 42.6 \\
    Khasi                    & Other/Small          & 717 & 21.3 & 38.1 \\
    Kiga                     & Niger-Congo          & 720 & 21.2 & 43.5 \\
    Sango                    & Creole               & 721 & 21.1 & 44.4 \\
    Zapotec                  & Other/Small          & 721 & 19.8 & 37.2 \\
    Nahuatl (E. Huasteca)    & Other/Small          & 721 & 19.7 & 38.4 \\
    Divehi                   & Indo-European        & 720 & 19.0 & 38.6 \\
    Nepalbhasa (Newari)      & Sino-Tibetan         & 721 & 18.4 & 35.9 \\
    Meadow Mari              & Other/Small          & 720 & 18.3 & 38.9 \\
    Twi (Akan)               & Niger-Congo          & 721 & 18.3 & 38.0 \\
    Rundi                    & Niger-Congo          & 719 & 17.5 & 37.4 \\
    Ganda (Luganda)          & Niger-Congo          & 720 & 17.1 & 39.0 \\
    Yucatec Maya             & Other/Small          & 721 & 17.1 & 36.6 \\
    Aymara                   & Other/Small          & 721 & 16.9 & 35.5 \\
    Chuvash                  & Turkic               & 721 & 16.9 & 41.5 \\
    Baluchi                  & Indo-European        & 721 & 16.8 & 36.5 \\
    Breton                   & Indo-European        & 721 & 16.8 & 34.4 \\
    Northern Sotho (Sepedi)  & Niger-Congo          & 720 & 16.4 & 34.9 \\
    Makassar                 & Austronesian         & 721 & 16.1 & 30.5 \\
    Kituba                   & Creole               & 720 & 15.6 & 44.2 \\
    Hawaiian                 & Austronesian         & 721 & 14.6 & 30.5 \\
    Tulu                     & Other/Small          & 721 & 14.3 & 31.6 \\
    Batak Toba               & Austronesian         & 721 & 14.1 & 26.1 \\
    Uyghur                   & Turkic               & 718 & 13.5 & 31.2 \\
    Samoan                   & Austronesian         & 721 & 13.3 & 28.4 \\
    Oromo                    & Other/Small          & 718 & 13.2 & 37.6 \\
    Lingala                  & Niger-Congo          & 721 & 12.9 & 31.3 \\
    Malagasy                 & Austronesian         & 720 & 12.6 & 27.2 \\
    Somali                   & Other/Small          & 720 & 12.2 & 28.5 \\
    Scots Gaelic             & Indo-European        & 721 & 12.2 & 24.7 \\
    Chichewa (Nyanja)        & Niger-Congo          & 721 & 12.2 & 29.5 \\
    Quechua                  & Other/Small          & 721 & 12.1 & 32.7 \\
    Crimean Tatar            & Turkic               & 721 & 11.5 & 31.9 \\
    Tok Pisin                & Creole               & 721 & 11.4 & 27.9 \\
    Hmong                    & Other/Small          & 720 & 11.2 & 26.8 \\
    Iloko                    & Austronesian         & 721 & 11.2 & 26.4 \\
    Pangasinan               & Austronesian         & 720 & 11.1 & 33.8 \\
    Odia (Oriya)             & Indo-European        & 721 & 11.0 & 26.5 \\
    Sundanese                & Austronesian         & 721 & 10.8 & 21.2 \\
    Kurdish (Kurmanji)       & Indo-European        & 721 & 10.8 & 26.1 \\
    Khmer                    & Other/Small          & 721 & 10.4 & 27.7 \\
    Tatar                    & Turkic               & 721 &  9.8 & 26.8 \\
    Sindhi                   & Indo-European        & 721 &  9.3 & 25.1 \\
    Myanmar (Burmese)        & Sino-Tibetan         & 721 &  9.2 & 28.7 \\
    Kapampangan              & Austronesian         & 721 &  9.0 & 22.6 \\
    Malayalam                & Other/Small          & 721 &  8.9 & 25.9 \\
    Balinese                 & Austronesian         & 721 &  8.9 & 26.6 \\
    Sinhala (Sinhalese)      & Indo-European        & 721 &  8.9 & 27.6 \\
    Welsh                    & Indo-European        & 721 &  8.5 & 21.8 \\
    Kyrgyz                   & Turkic               & 721 &  8.3 & 24.8 \\
    Tetum                    & Austronesian         & 721 &  8.3 & 24.7 \\
    Betawi                   & Creole               & 721 &  8.2 & 18.2 \\
    Bikol                    & Austronesian         & 721 &  8.0 & 23.3 \\
    Buryat                   & Other/Small          & 720 &  7.9 & 31.0 \\
    Assamese                 & Indo-European        & 721 &  7.9 & 25.0 \\
    Javanese                 & Austronesian         & 721 &  7.4 & 21.6 \\
    Hiligaynon               & Austronesian         & 721 &  6.9 & 23.3 \\
    Gujarati                 & Indo-European        & 721 &  6.9 & 21.8 \\
    Telugu                   & Other/Small          & 721 &  6.8 & 22.2 \\
    Azerbaijani              & Turkic               & 721 &  6.5 & 20.9 \\
    Krio                     & Creole               & 721 &  6.4 & 21.9 \\
    Cantonese                & Sino-Tibetan         & 721 &  6.1 & 17.8 \\
    Mongolian                & Other/Small          & 721 &  5.8 & 25.5 \\
    Nepali                   & Indo-European        & 721 &  5.3 & 21.9 \\
    Awadhi                   & Indo-European        & 721 &  5.0 & 21.8 \\
    Bhojpuri                 & Indo-European        & 721 &  4.9 & 20.5 \\
    Pashto                   & Indo-European        & 721 &  4.7 & 24.0 \\
\end{longtable}

\section{Prompt and Language Coverage}
\label{app:appendixJ}

\begin{table}[H]
  \caption{Unique prompt counts by category.}
  \label{tab:prompts}
  \centering
  \begin{tabular}{lc}
    \toprule
    \textbf{Category} & \textbf{Harmful} \\
    \midrule
    Dangerous Behavior \& Criminal Content & 35 \\
    Hate \& Discrimination                 & 20 \\
    Violence \& Threats                    & 18 \\
    Self-Harm \& Suicide                   & 15 \\
    Misinformation or Disinformation       & 12 \\
    Other                                  & 9  \\
    Explicit Sexual Content                & 3  \\
    Model Security                         & 3  \\
    Sensitive Personal Information \& PII  & 2  \\
    Crimes Against Children                & 1  \\
    Sexual Violence \& Illegal Sexual Acts & 1  \\
    \midrule
    \textbf{Total}                         & \textbf{119} \\
    \bottomrule
  \end{tabular}
\end{table}

\begin{table}[H]
  \caption{Count of low-resource languages by family. Rows listing multiple families indicate ties in count. Coverage is concentrated: two families hold 30 of the 78 languages, while 13 families contribute three or fewer.}
  \label{tab:families}
  \centering
  \begin{tabular}{p{0.72\linewidth}c}
    \toprule
    \textbf{Language Family} & \textbf{Count} \\
    \midrule
    Austronesian; Indo-European                                          & 15 \\
    Niger-Congo                                                          & 9  \\
    Turkic                                                               & 6  \\
    Creole                                                               & 5  \\
    Sino-Tibetan; Nilo-Saharan                                           & 4  \\
    Dravidian; Mayan                                                     & 3  \\
    Mongolic; Afro-Asiatic; Austro-Asiatic                               & 2  \\
    Quechuan; Aymaran; Uralic; Hmong-Mien; Otomanguean;
    Uto-Aztecan; Abkhaz-Adyghe; Kra-Dai                                 & 1  \\
    \bottomrule
  \end{tabular}
\end{table}

\end{document}